\documentclass[runningheads]{llncs}
\usepackage{mathrsfs}
\usepackage[utf8]{inputenc}
\usepackage{amssymb}
\usepackage{caption}
\usepackage{amsmath}
\usepackage{cite}
\usepackage{subcaption}
\usepackage{todonotes}
\usepackage[mode=buildnew]{standalone}%
\usepackage{microtype}
\usepackage{verbatim}
\usepackage{booktabs}
\usepackage{siunitx}
\usepackage{bm}
\usepackage[hyphens]{url}
\usepackage[pagebackref=true,breaklinks=true]{hyperref}
\hypersetup{
    colorlinks,
    linkcolor={red!75!black},
    citecolor={blue!75!black},
    urlcolor={blue!75!black},
}

\usepackage[capitalise]{cleveref}

\usepackage{tikz}
\usepackage{pgfplots}
\usetikzlibrary{spy,calc,positioning,patterns,fit,calc,shapes}

\usepackage[inline]{enumitem}
\setlist*[enumerate]{label=(\arabic*)}

\usepackage{siunitx}
\usepackage{import}
\usepackage{xspace}
\makeatletter
\DeclareRobustCommand\onedot{\futurelet\@let@token\@onedot}
\newcommand{\@onedot}{\ifx\@let@token.\else.\null\fi\xspace}
\makeatother

\newcommand{\etal}[1]{#1 \emph{et~al\onedot}}

\newcommand{\eg}{e.\,g.,\xspace}
\newcommand{\vs}{vs\onedot}

\definecolor{faublue}{RGB}{0,51,102}
\definecolor{color_ours}{RGB}{253,174,97}
\definecolor{color_GANwriting}{RGB}{255,255,191}
\definecolor{color_lineGAN}{RGB}{171,217,233}
\definecolor{color_real}{RGB}{44,123,182}

\usepackage[toc,page]{appendix}

\newenvironment{customlegend}[1][]{%
	\begingroup
	\csname pgfplots@init@cleared@structures\endcsname
	\pgfplotsset{#1}%
}{%
	\csname pgfplots@createlegend\endcsname
	\endgroup
}%

\def\addlegendimage{\csname pgfplots@addlegendimage\endcsname}
\newcommand{\addlegendimageintext}[1]{%
	\tikz {
		\begin{customlegend}[
			legend entries={\empty},
			legend style={
				draw=none,
				inner sep=0pt,
				column sep=0pt,
				nodes={inner sep=0pt}}]
			\addlegendimage{#1}
		\end{customlegend}
	}%
}

\begin{document}
    \title{SmartPatch: Improving Handwritten Word Imitation with Patch Discriminators}
    \author{
    Alexander Mattick\orcidID{0000-0001-7805-199X} \and
    Martin Mayr\orcidID{0000-0002-3706-285X} \and
    Mathias Seuret\orcidID{0000-0001-9153-1031} \and
    Andreas Maier\orcidID{0000-0002-9550-5284} \and
    Vincent Christlein\orcidID{0000-0003-0455-3799}
    }
    \authorrunning{A. Mattick et al.}
    \titlerunning{SmartPatch}
    \institute{ 
    Pattern Recognition Lab, \\Friedrich-Alexander University Erlangen-Nürnberg, Germany
    }
	\maketitle
	\begin{abstract}
	    As of recent generative adversarial networks have allowed for big leaps in the realism of generated images in diverse domains, not the least of which being handwritten text generation. 
	    The generation of realistic-looking hand\-written text is important because it
	    can be used for data augmentation in handwritten text recognition (HTR) systems or human-computer interaction. 
	    We propose SmartPatch, a new technique increasing the performance of current state-of-the-art methods by augmenting the training feedback with a tailored solution to mitigate pen-level artifacts.
	    We combine the well-known patch loss with information gathered from the parallel trained handwritten text recognition system and the separate characters of the word.
	    This leads to a more enhanced local discriminator and results in 
	    more realistic and higher-quality generated handwritten words.
	\end{abstract}

	\begin{figure}
	    \centering
	    \captionsetup[subfigure]{labelformat=empty}
	    \begin{tikzpicture}
	        \begin{scope}[spy using outlines={circle,lens={scale=4},height=2cm,width=2cm, connect spies, ultra thick}, node distance=1cm]
	        \node[](base1){%
	            \includegraphics[height=28px]{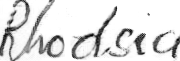}
	        };
	        \node[right=\textwidth of base1.west, anchor=east](character1){%
	            \includegraphics[height=28px]{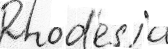}
	        };
	        \spy[orange,thick] on ($(base1)+(-1.3cm,0.23cm)$) in node (s1) [right=0.5cm of
            base1.north east,anchor=north west];
            \spy[orange,thick] on ($(character1)+(-1.6cm,0.15cm)$) in node (s2) [left=0.5cm of
            character1.north west,anchor=north east];
            \end{scope}
            \draw[->,ultra thick, shorten >=1ex, shorten <=1ex](s1)--(s2);
	    \end{tikzpicture}
	    \caption{Reduction of pen-level artifacts with tailored patch discriminator.}
	    \label{fig:teaser}
	\end{figure}

	\section{Introduction}
    Automatically generating handwritten text is an active field of research due to its high complexity and due to multiple possible use cases.
    Especially for handwritten text recognition (HTR) of historical documents, the possibility to produce extra writer-specific synthetic training data could be a game-changer, because manual transcriptions of a sufficient amount of epoch and document-specific data can be very cost- and time-intensive for research projects dealing with ancient documents.
    Synthetically produced data that are based on the type of document and styles of the different writers can drastically improve the performance of HTR systems.
    
	Algorithms for handwriting generation can be divided into two different categories: online generation, in which the dataset contains strokes as a time-series, and offline generation, in which only the result of the text is accessible. 
	While online generation has the opportunity to exploit the sequential nature of real handwritten texts, it has the drawback of the complicated assessment of data. 
	In contrast, there are quite many ways to get images of text lines for an offline generation.

	The state-of-the-art offline handwriting generation model on word-level, known as GANwriting~\cite{kang_ganwriting_2020} suffers from unrealistic artifacts. This greatly reduces the degree of authenticity of the generated handwriting and often allows quick identification.
	In this paper, we extend their approach by adding a patch level discriminator to mitigate often produced artifacts. 
	We are going to introduce three different variations: a naive version (NaivePatch) that follows a sliding-window approach, a centered version (CenteredPatch) which uses the attention window of the jointly-trained recognition system, and a smart version (SmartPatch) that additionally inputs the expected character into the patch discriminator.
	In particular, we make the following contributions: 
    \begin{enumerate}
        \item We extend the current state-of-the-art method for handwritten word imitation~\cite{kang_ganwriting_2020} with an extra lightweight patch-based discriminator. 
        We derive this additional discriminator from a naive version, via a patch-centered version that incorporates attention masks from the handwritten word recognition system, and finally, propose a new smart discriminator that also uses a patch-centered version of the word images in combination with knowledge of the characters' positions and the recognizer's predictions.\footnote{Code: \url{https://github.com/MattAlexMiracle/SmartPatch}}
        \item We further show that HTR can robustly validate the synthetically produced outputs. 
        \item Finally, we thoroughly evaluate our results compared to other methods and also to genuine word images through qualitative and quantitative measurements.
    \end{enumerate}

	This work is structured as follows.
	First, we give an overview of current related work in \cref{related_work}.
	The methodology is outlined in \cref{methodology}.
    We describe and analyze GANwriting's method~\cite{kang_ganwriting_2020}, which serves as the basis for our approach. 
	We present our extension to improve robustness against pen-level artifacts. 
	In \cref{evaluation}, we evaluate our approach and compare the produced outputs against other methods and real data. 
	First, we compare the FID score of the approaches. 
	Second, we measure the readability with the usage of a sequence-to-sequence handwritten word recognition system. 
	And third, a user study is conducted to assess which data is looking most realistic to human subjects. 
	Finally, we discuss our results in \cref{discussion} and conclude the paper with \cref{conclusion}.
	
	\section{Related Work}
	\label{related_work}
	Previous approaches to generating human handwriting can be divided by their data used for training the models: online and offline handwriting. 
	\paragraph{Online Handwriting Synthesis.}
	The first method which achieved convincing-looking outputs was the method of Alex Graves~\cite{graves_generating_2014}. This approach is based on Long Short-Term Memory (LSTM) cells to predict new positions given the previous ones. Also, skip connections and an attention mechanism were introduced to improve the results.
	The main problem of this method is the often changing writing style towards the end of the sequence.
	Chung et al.~\cite{chung15vrnn} overcome this problem by using Conditional Variational RNNs (CVRNNs).
	Deepwriting~\cite{aksan_deepwriting_2018} separately processed \emph{content} and \emph{style}. 
	\emph{Content} in this case describes the information/words to be written, while \emph{style} describes writer-specific glyph attributes like slant, cursive \vs non-cursive writing, and variance between glyphs.
	The disadvantage of this method is the necessary extra information in the online handwriting training data.
	Beginning and ending tokens of every single character have to be added to the data to produce persistent results. 
	Aksan et al.~\cite{aksan2018stcn} introduced Stochastic Temporal CNNs replacing the CVRNNs in the architecture to further reduce inconsistency in generating realistic-looking handwriting.
	Another approach for generating online handwriting \cite{bhattacharya_sigma-lognormal_2017} uses the theory of kinematics to generate parameters for a sigma-lognormal model. These parameters may later be distorted to produce new synthetic samples.
	The problem of all these methods is the crucial need for online handwriting data for training and also for inference. 
	Mayr et al.~\cite{mayr_spatio-temporal_2020} solved this for inference by converting offline data into online handwriting, processing the online data, and going back to the spatial domain. Still online data is necessary for training the writer style transfer. 
	Another disadvantage of this method is the lack of an end-to-end approach, changing one stage in the pipeline leads to the retraining of all following stages.
	\paragraph{Offline Handwriting Synthesis.}
    Currently, many offline handwriting generation methods make use of the generative adversarial network paradigm~\cite{goodfellow2014generative} to simultaneously generate either words or entire lines. ScrabbleGAN~\cite{fogel_scrabblegan_2020} uses stencil-like overlapping generators to build long words with, while consistent, random style. A big advantage of ScrabbleGAN is the use of unsupervised learning which allows for many more real-world use cases due to the increase of available training data. GANwriting~\cite{kang_ganwriting_2020} can be seen as development on top of ScrabbleGAN without the stencil-like generators, but with the ability to select specific styles based on author examples. Davis et al.~\cite{davis2020text} present further development by using a mixture of GAN and autoencoder training to generate entire lines of handwritten text based on the extracted style. This approach shares some similarities to ours in the sense that they also use the recognizer's attention window to localize characters within the image, though they use this for style extraction rather than improved discriminators. For convenience, we further call the method of Davis et al. lineGen.

	\section{Methodology}
	\label{methodology}
	\begin{figure}[t]
	    \centering
	    \includegraphics[width=0.95\textwidth]{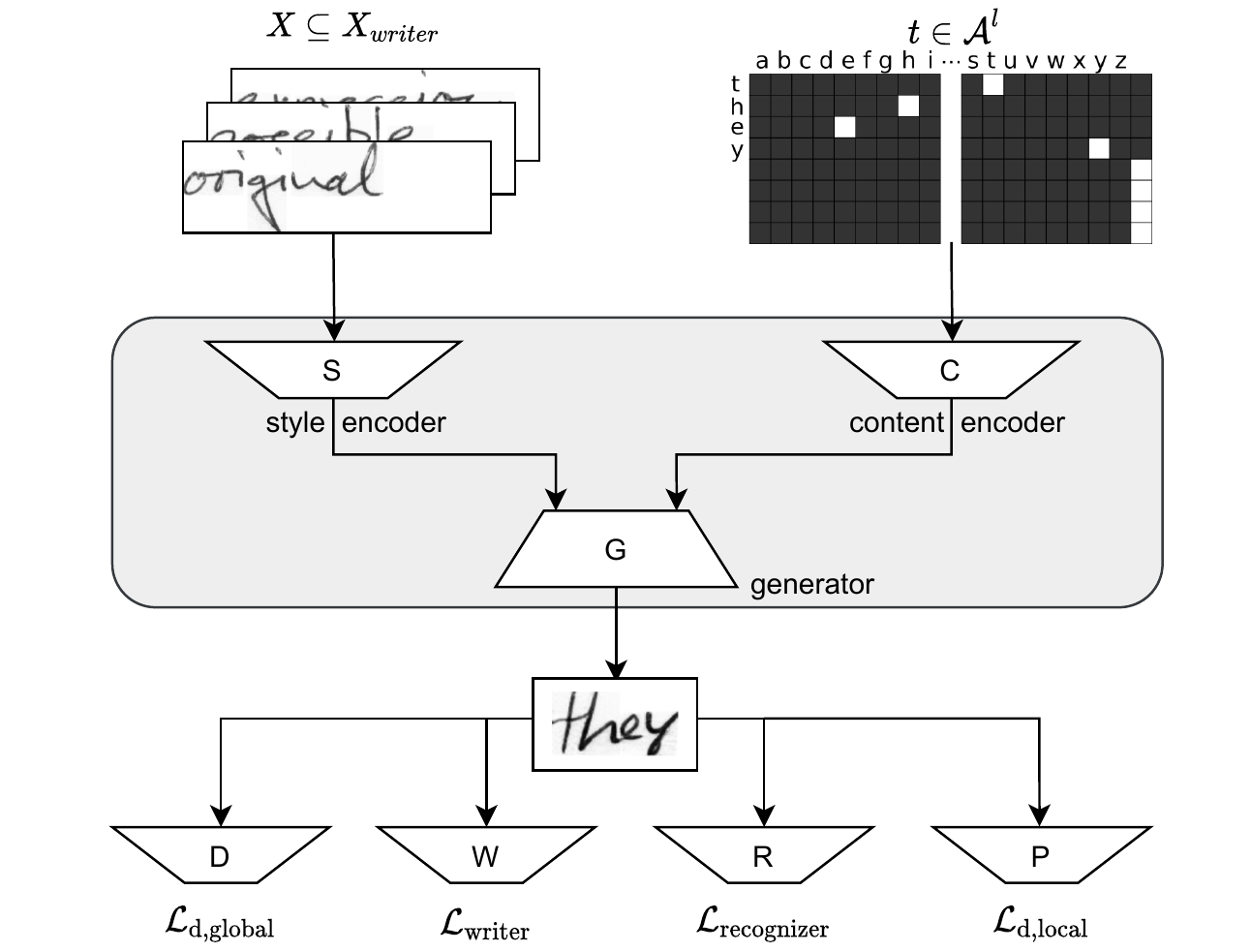}
	    \caption{Basic architecture sketch. Given an input image and new text, a GAN-based approach with additional loss terms generates the text with the style of a selected writer. 
	    }
	    \label{fig:architecture}
	\end{figure}
	\subsection{Background}
	\Cref{fig:architecture} depicts our architecture. In large parts, we follow the work of GANwriting~\cite{kang_ganwriting_2020} but additionally introduce a local patch loss.  %
	It consists of six interconnected sub-models. The style encoder encodes several example words obtained by an author into a corresponding style-latent space.
	The content encoder transforms a one-hot encoded character-vector into two content vectors $g_1(c)$ and $g_2(c)$ where $g_1(c)$ works on an individual character level and is concatenated with the style-vector. Conversely, $g_2(c)$ encodes the entire input string into a set of vectors, which later serves as input for the adaptive instance normalization~\cite{AdaIn} layers in the decoder. 
	The concatenated content-style vector is then used in the fully-convolutional generator. 
	This generator consists of several %
	AdaIn layers~\cite{AdaIn} which injects the global content vector $g_2(c)$ to different up-sampling stages. 
	The original learning objective contains three units:
	\begin{enumerate*}
	    \item A writer classifier that tries to predict the style encoder's input scribe, 
	    \item a handwritten text recognizer (specifically~\cite{textRecognition}) consisting of a VGG-19 backbone~\cite{simonyan2015deep}, an attention-layer, and a two-layered bidirectional GRU~\cite{cho2014gru}, 
	and 
	\item a discriminator model. 
	\end{enumerate*}
	The text and writer classifier are both trained in a standard supervised manner, while the discriminator is trained against the generator following the generative-adversarial network~\cite{goodfellow2014generative} paradigm: The discriminator is trained to maximize the difference between real and fake samples, while the generator is trained to minimize it.  
	The total generator loss then consists of a sum of the losses of text, writer, and discriminator network~\cite{kang_ganwriting_2020}:
	\begin{align}\label{eq:GANwriting}
        \min_{G,W,R}\max_D \mathcal{L}(G, D, W, R) = \mathcal{L}_\text{d}(G,D) + \mathcal{L}_\text{w}(G,W)+ \mathcal{L}_\text{r}(G,R),
	\end{align}
	where $G$ is the text generated by the generator, $D$ is real data, $W$ is the real writer and $R$ the real text input into the encoder.
	$\mathcal{L}_\text{d}, \mathcal{L}_\text{w}, \mathcal{L}_\text{r}$ are the discriminator, writer, and recognizer loss respectively.

	\begin{figure}[t]
	    \centering
	    \captionsetup[subfigure]{labelformat=empty}
	    \begin{tikzpicture}
	        \begin{scope}[spy using outlines={rectangle,lens={scale=3},height=1.5cm,width=1.5cm, connect spies, ultra thick}, node distance=1cm]
	        \node[](base1){
	            \includegraphics[height=28px]{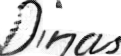}
	        };
	        \node[right=\textwidth of base1.south west, anchor=south east](character1){%
	            \includegraphics[height=28px]{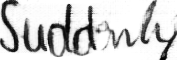}
	        };
            \spy[orange,thick] on ($(base1)+(-0.8cm,0.2cm)$) in node (s1) [right=0.5cm of
            base1.north east,anchor=north west];
            \spy[orange,thick] on ($(character1)+(0.2cm,-0.1cm)$) in node (s2) [left=0.5cm of
            character1.north west,anchor=north east];
	        \end{scope}
	    \end{tikzpicture}
	    \caption{Frequently appearing artifacts in the outputs of GANwriting.}
	    \label{fig:bandingInGANwriting}
	\end{figure}
	
	Even though GANwriting produces sufficiently accurate words at a glance, closer inspection of the generated words shows some typical artifacts visualized in \cref{fig:bandingInGANwriting}.
	The left image shows noticeable vertical banding artifacts for the letter ``D''.
	It is quite unlikely to see such a pattern in real images because humans tend to produce continuous strokes in their handwriting.
	Another prominent artifact is depicted on the right-hand side. 
	There the letter ``e'' in the middle of the word almost vanishes. 
	By contrast, the strokes in real handwriting have regular intensities throughout the word, except the pen runs out of ink which could be the case for the last letters, but not only for one letter in the middle of the word.
	We hypothesize that the flaws in the pen-level happen due to the global discriminator loss not being fine-grained enough to discourage locally bad behavior. Therefore, we introduce additional local losses to explicitly discourage unrealistic line appearance.
	
	\subsection{Local discriminator design}
	
	\paragraph{Naive patch discriminator (NaivePatch).}
	We first determine the baseline-efficacy of using local discriminators, by naively splitting the image into overlapping square patches of size height$\times$height. We use a step size of $\frac{\text{height}}{2}$ making sure that there's considerable overlap between patches as to not generate hard borders between patch zones. A similar strategy in the generator has already been proven to be advantageous by ScrabbleGAN~\cite{fogel_scrabblegan_2020} which learned individual overlapping ``stencil'' generators for each glyph. The expected local loss is then added to \cref{eq:GANwriting}. Let $q\in P_\text{real}$, $p\in P_\text{fake}$ be the set of patches stemming from image $D$,$G$, resulting in the new loss term
	\begin{equation}
        \min_{G,W,R}\max_D \mathcal{L}(G, D, W, R) = \mathcal{L}_{\text{d,glob}}(G,D) + \mathcal{L}_\text{w}(G,W)+ \mathcal{L}_r(G,R)+ \mathcal{L}_{\text{d,loc}}(p,q).
	\end{equation}
	The local discriminator model was chosen as small as possible to benchmark the addition of local discriminators, rather than simply the scaling of GANwriting with additional parameters. Specifically, we selected the discriminator design introduced in the Pix2Pix architecture~\cite{isola2018imagetoimage}. 
	The receptive field was chosen to be \num{70x70}, as this was the smallest size that entirely covers the patch. This discriminator design shows to
	be the right size to work with our small patches without over-parameterizing the discriminator.
	In contrast to PatchGAN~\cite{isola2018imagetoimage} we use this discriminator only on the small patches extracted from the image, rather than convolving the discriminator across the whole image like PatchGAN.
	The discriminator loss function is the standard saturating loss presented in~\cite{goodfellow2014generative} applied to the patches $p\in P_\text{fake},q\in P_\text{real}$ resulting in
	\begin{equation}
	    \mathcal{L}_{\text{d,loc}}(p,q) = \mathbb{E}[\log D_{\text{loc}}(q)] + \mathbb{E}[1-\log D_{\text{loc}}(p)].
	\end{equation}
	
	\paragraph{Centered patch discriminator (CenteredPatch).}

	\begin{figure}[t]
	\captionsetup[subfigure]{labelformat=empty}
    \begin{subfigure}[t]{.33\textwidth}
    \centering 
      \includegraphics[width=.62\linewidth]{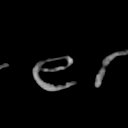}
      \caption{Patch centered at ``e''}
      \label{fig:sfig1}
    \end{subfigure}%
    \begin{subfigure}[t]{.33\textwidth}
    \centering 
      \includegraphics[width=.62\linewidth]{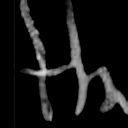}
      \caption{Patch centered at ``H''}
      \label{fig:sfig2}
    \end{subfigure}%
    \begin{subfigure}[t]{.33\textwidth}
    \centering 
      \includegraphics[width=.62\linewidth]{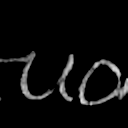}
      \caption{Patch centered at ``u''}
      \label{fig:sfig3}
    \end{subfigure}
    \caption{Centered patches around letters, automatically found using text recognition module's attention windows.}
    \label{fig:centered_characters}
    \end{figure}
	We further explore the idea of using the text recognition system trained alongside the generator as a useful prior for locating sensible patches in the samples: As part of the sequence-to-sequence model used for digit recognition, an attention map is generated across the latent space of the generator~\cite{textRecognition}. Since the encoder up to this point is fully convolutional, we can assume similar locality in embedding as in feature space. This means that by up-sampling the attention-vector to the full width of the word image, we obtain windows centered on the associated character. This also allows for a flexible number of patches per image. We found linear up-sampling and the use of the maximum of the attention window as the center led to the most accurate location around specific letters (see \cref{fig:centered_characters}). 
	Simple nearest-neighbor up-sampling or matching the relative position of the maximum value in the attention window and the width of the image did not produce consistent results, presumably because different letters set their maximal attention-window at slightly different points relative to the letter's center. Preliminary tests with adapting the patch size depending on character proved very unstable in early training and have not been analyzed further.
	The loss function, patch size, and model are the same as in NaivePatch to maintain comparability between approaches.
	
	\paragraph{Smart patch discriminator (SmartPatch).}
	\begin{figure}[t]
	    \centering
	    \includegraphics[width=1\textwidth]{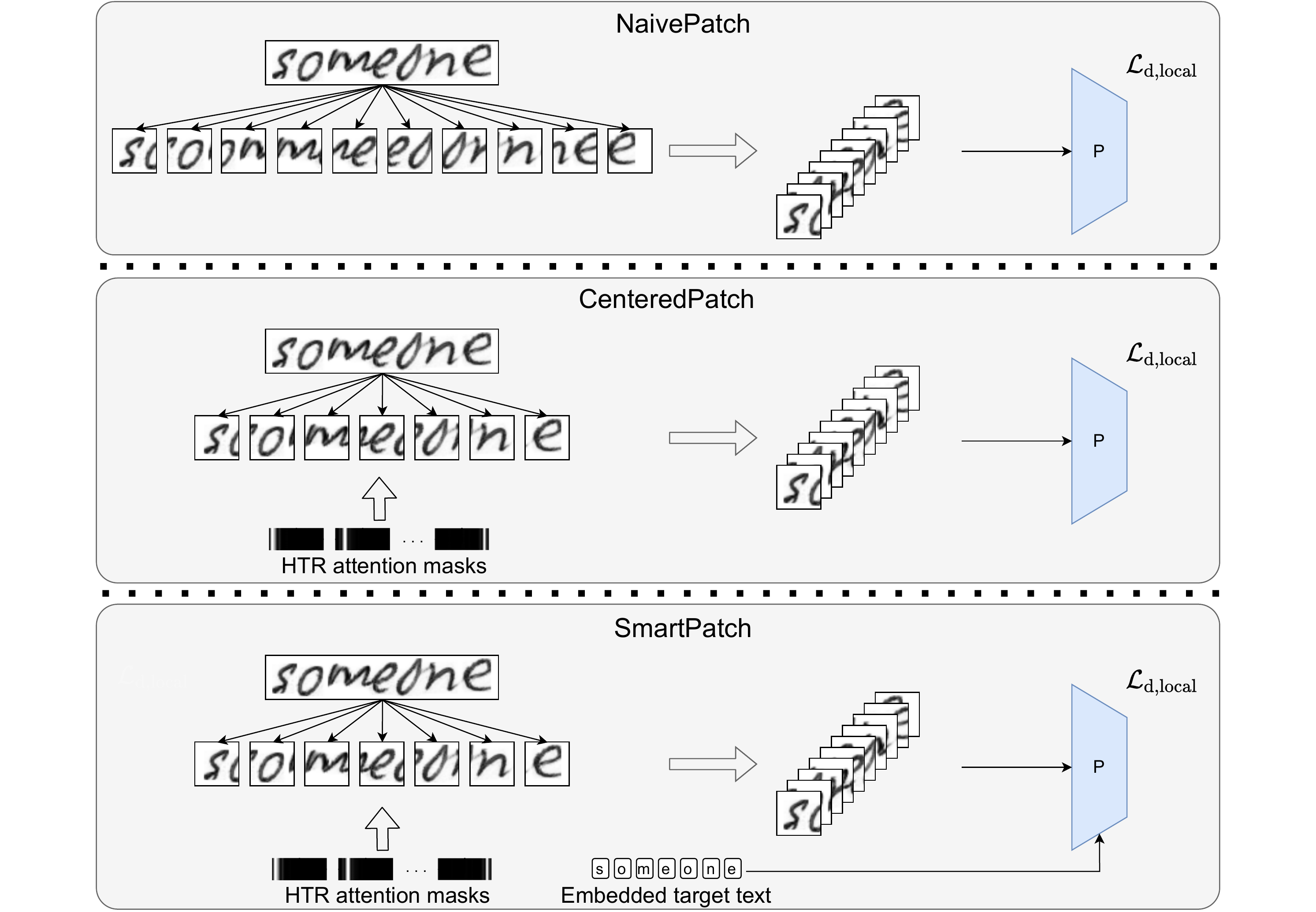}
    	\caption{Architectural overview of our methods: All discriminator architectures share the same Pix2Pix discriminator design\cite{isola2018imagetoimage}. In the case of the smart patch discriminator, we use a linear layer to project the encoded text into a compact latent-space and inject this as additional channels before the discriminator's second-to-last layer. The latter two approaches use the recognition module's up-sampled attention masks for accurately localizing the center of patches.}
	\end{figure}
		\begin{figure}
	    \centering
	    \includestandalone[width=\linewidth]{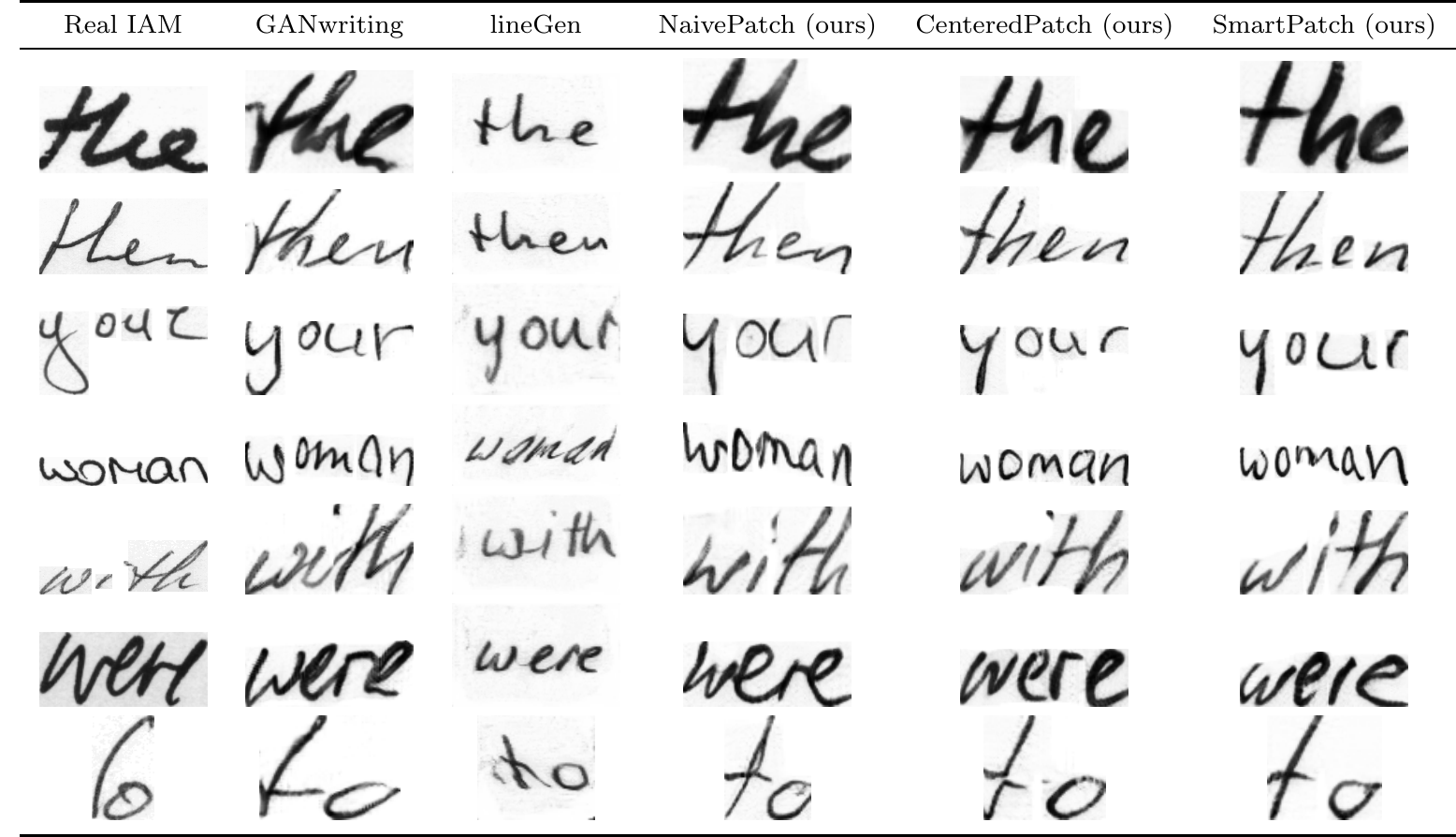}
	    \caption{Randomly chosen outputs of different words. For each row the priming image and the content is the same. More examples can be found in appendix \ref{appendix:examples}.}
	    \label{fig:results}
	\end{figure}
	The centered patch discriminator (CenteredPatch) is extended further by injecting the label of the expected character into the patch.
	This is done by first projecting the one-hot encoded character vector into a 
	compact latent space $C_\text{enc}\in\mathbb{R}^8$ and then appending the vector as extra channels into the CNN's penultimate layer.
	In this way, we minimize the architectural differences between the NaivePatch, CenteredPatch and SmartPatch.
	\begin{equation}
	    \mathcal{L}_{\text{d,loc}}(p,q,c_{\text{real}},c_{\text{fake}}) = \mathop{\mathbb{E}}[\log D_{\text{loc}}(q,c_{\text{real}})]-\mathop{\mathbb{E}}[1-\log D_{\text{loc}}(p,c_{\text{fake}})]
	\end{equation}
	where $c_\text{real}\in C_\text{enc}$ and $c_\text{fake}\in C_\text{enc}$ are the character-labels for the fake and real image, respectively. In the case of generated images, we also use the actual character that is supposed to be at the given location, rather than the prediction supplied by the recognition module. This stabilizes training especially for writers with bad handwriting where the recognition module doesn't produce good predictions.

	\section{Evaluation}
	\label{evaluation}
	\setlength{\tabcolsep}{12pt}
	\begin{table}[t]
	    \centering
	    \caption{Fr\'echet Inception Distance (FID) computed between real data and the synthetic data. Also, Character-Error-Rate (CER) and loss of the handwritten text recognition system (HTR loss). All approaches using local discriminators seem to perform significantly better than the baseline. We also want to highlight the lower run-to-run variance for the patch discriminators.}
    	\begin{tabular}{lrrr}
    	    \toprule
    	    Architecture & FID & CER & HTR loss\\
    	    \midrule
    	    Real IAM               & -- &  $6.27\pm 0.00$       & $23.68\pm 0.00$\\
    	    GANwriting\cite{kang_ganwriting_2020} & $51.16$ &  $4.68\pm 0.54$& $17.02\pm 0.17$ \\
    	    NaivePatch (ours)           & $\textbf{40.33}$ &  $4.32\pm 0.33$& $13.00\pm 0.12$ \\
    	    CenteredPatch (ours)           & $42.05$ &  $4.18 \pm 0.24$& $13.41\pm 0.05$\\
    	    SmartPatch (ours)       & $49.00$ &  $\textbf{4.04}\pm 0.18$& $\textbf{12.54}\pm 0.06$\\
    	    \bottomrule
    	\end{tabular}
    	\label{table:quantitative}
	\end{table}
	\setlength{\tabcolsep}{6pt}
	\begin{figure}[t]
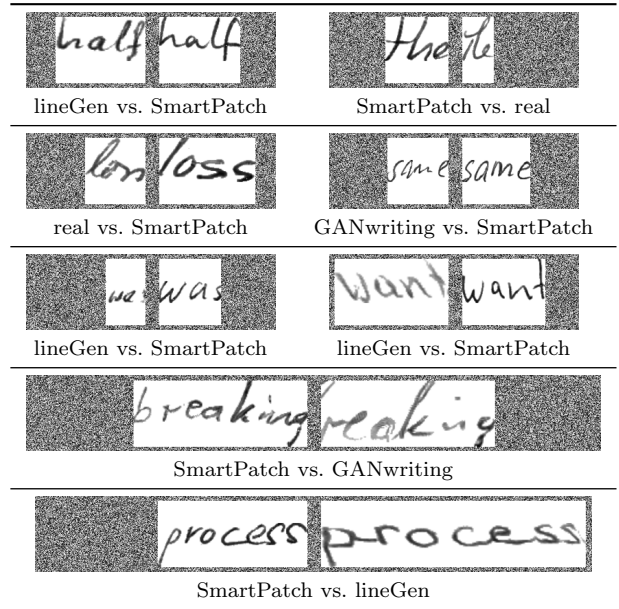

	    \centering
	    \includestandalone{us_examples}
	    \caption{Examples of the user study including real data and synthetic data produced by GANwriting, lineGen, and our approach SmartPatch.}
	    \label{fig:Userstudy}
    \end{figure}
	We compare the quality of the generated words using three evaluation metrics. 
	First, we make use of the  Fréchet Inception Distance (FID)~\cite{heusel2018gans} to measure the feature similarity. This metric is widely used in the field of generation tasks in combination with GANs. 
	Second, we measure the legibility of our synthetically created data with a word-based HTR system trained on real data.
	Third, we conduct a user study to see if we can robustly forge handwriting that humans don't recognize the difference.
	
	For testing, we re-generate data of the IAM dataset~\cite{marti02iam} using the RWTH Aachen test split. Some random examples can be seen in \cref{fig:results}.
	For the \emph{style decoder}, we make sure to shuffle each writer's sample images after each generated word to facilitate the generation of different instances of words with the same content. In this way, we avoid creating the image for the same writer and text-piece multiple times and to give us a more accurate assessment of the mean performance of each of the approaches. Aside from removing horizontal whitespaces from small words (as is done in the IAM dataset itself), we perform no additional augmentation or alteration of the GAN outputs.
	
	\subsection{FID Score}
	\cref{table:quantitative} shows the FID score between the real data and the synthetic samples. 
	Empirically, all our three proposed methods produce a lower FID score than GANwriting~\cite{kang_ganwriting_2020} with the naive approach ranking best. 
	However, note that these values should be taken with a grain of salt: FID is computed using an Inception Net~\cite{szegedy2015rethinking} trained on ImageNet~\cite{russakovsky2015imagenet}. Empirically, this produces sound embeddings for generative models trained on ImageNet or similar datasets, but not necessarily for domains far away from ImageNet, such as monochromatic handwritten text lines. This has also been noted by \etal{Davis}~\cite{davis2020text}. There is also the problem of how the conditional generation capabilities of handwriting generation is used: Should you use random content? Should you choose random noise for the style? We opted to use the same ``synthetic IAM'' approach as used later in the HTR tests as this should provide a lower bound on unconditional generation as well (with random content and text the distributions will diverge more). In general, we mainly report FID for comparability with GANwriting~\cite{kang_ganwriting_2020}. %
	
	\subsection{Handwritten Text Recognition}\label{handwritten Test}
	We evaluate the legibility of the words by training a state-of-the-art handwritten text recognition model~\cite{kang_convolve_2019} on the IAM dataset~\cite{marti02iam} and by performing evaluations on synthetic data from the baseline, as well as on all our three patch discriminator outputs. For training the text recognition model, we use the RWTH Aachen partition and remove all words containing non-Latin symbols (such as ``\,!\,.\,?\,)\,'') and all words with more than ten characters, as neither the baseline nor our approaches were trained using long words or special characters.
	
	We chose the HTR system by \etal{Kang}~\cite{kang_convolve_2019} due to its high performance and because it is different from the one used in the adversarial loss of the generation systems. In this way, we account for potential biases introduced in the generator's architecture.
	
	We observe that approaches augmenting the global loss with local discriminators produce more (machine)-legible content, with more sophisticated approaches further reducing the CER shown in \cref{table:quantitative}. We especially want to highlight the stability of the CenteredPatch, and SmartPatch: We observe significantly less variance across evaluation runs, which lends credence to our claim that locally attending discriminators to symbols allows for the reduction of extreme failure cases (this can also be observed when visually inspecting the samples).

	\subsection{User Study}

	\begin{figure}
	    \centering
	    \begin{tikzpicture}
			\begin{customlegend}[%
				legend entries={Real data, GANwriting, lineGen, SmartPatch (ours)},
				legend style={%
					/tikz/every even column/.append	style={column sep=0.5cm}},
				legend columns=-1,
				legend cell align=left]
				\addlegendimage{draw=faublue,fill=color_real,area legend}
				\addlegendimage{draw=faublue,fill=color_GANwriting,area legend}
				\addlegendimage{draw=faublue,fill=color_lineGAN,area legend}
				\addlegendimage{draw=faublue,fill=color_ours,area legend}
			\end{customlegend}
		\end{tikzpicture}

        \vspace{2pt}
		\begin{tikzpicture}
			\begin{axis}[%
				ymax=100,
				ymin=0,
				ymajorgrids,
				ylabel={Pick rate [\%]},
				width=0.95\textwidth,
				height=0.3\textheight,
				xtick pos=left, %
				symbolic x coords={a,b,c,d,e,f,g,h,i,j},
				xtick={a,b,c,d,e,f,g,h,i,j},
				ytick={25,50,75},
				x tick label style={yshift={-0.5em-mod(\ticknum,2)*1em}},
				xticklabels={\scriptsize SmartPatch \vs GANwriting, 
				\scriptsize SmartPatch \vs lineGen, 
				\scriptsize SmartPatch \vs Real, 
				\scriptsize GANwriting \vs Real, 
				\scriptsize lineGen \vs Real, 
				\scriptsize lineGen \vs GANwriting},
				xticklabel style={align=center},
				bar width=15pt,
				minor y tick num=0,
				every node near coord/.append style={
					/pgf/number format/fixed zerofill,
					/pgf/number format/precision=1
				},
				ybar stacked,
				]
				\addplot[ybar,black,fill=color_ours] coordinates 
				{(a,64.5) (b,70.5) (c,54.4) (d,0) (e,0) (f,0) };
				\addplot[ybar,black,fill=color_lineGAN] coordinates
				{(a,0) (b,29.5) (c,0) (d,0) (e,28.1) (f,46.5)};
				\addplot[ybar,black,fill=color_GANwriting] coordinates
				{(a,35.5) (b,0) (c,0) (d,34.0) (e,0) (f,53.5)};
				\addplot[ybar,black,fill=color_real] coordinates
				{(a,0) (b,0) (c,45.6) (d,66.0) (e,71.9) (f,0)};
				\coordinate (A) at (axis cs:a,50);
                \coordinate (O1) at (rel axis cs:0,0);
                \coordinate (O2) at (rel axis cs:1,0);
			\end{axis}
			                \draw [black,sharp plot,dashed] (A -| O1) -- (A -| O2);
		\end{tikzpicture}
		\caption{Pick rate in the user study. Each bar represents one comparison.}
	    \label{fig:us_stacked}
	\end{figure}
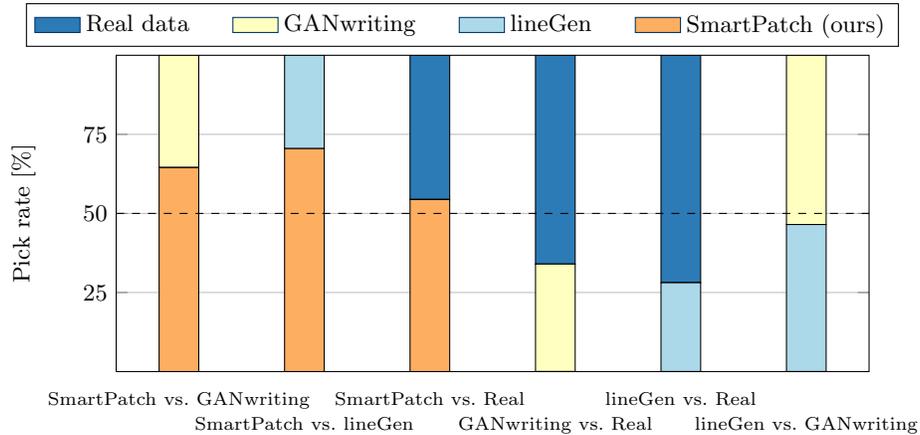
	In order to properly assess the quality of our handwriting system, we conducted a user study~\footnote{\url{https://forms.gle/TNoZvxihJNUJiV1b9}} comparing the produced synthetic words of GANwriting~\cite{kang_ganwriting_2020}, handwritten line generation~\cite{davis2020text}, and our best working method SmartPatch against each other. We also compare each approach against real images of the test set of IAM~\cite{marti02iam}. 
	Example samples can be seen in \cref{fig:Userstudy}.

	Participants were instructed to determine the more realistic sample, given two images from different sources. The samples were created by sampling uniformly at random from the synthetic IAM dataset described in \cref{handwritten Test}, and matching pairs of samples with the same content and writer. For line generation (lineGen), we used the code and models provided by~\cite{davis2020text} and generated individual words from the IAM dataset. Since it was not trained on individual words, but rather on entire lines, 
	the comparison between GANwriting and our method is not perfectly fair, but we choose to include it in our benchmark as it is another strong baseline to compare against. All samples were normalized and aligned to a common baseline, with the background being uniform noise to cancel out any decision-impacting bias, such as contrast.
	
	In total, the study consists of six sets with twenty images each, comparing the configurations SmartPatch \vs GANwriting, SmartPatch \vs lineGen, SmartPatch \vs Real, Real \vs GANwriting, Real \vs lineGen and GANwriting \vs lineGen. 
		
	\cref{fig:us_stacked} shows the pick rate for each comparison in percent.
	The pick rate is defined as the percentage one image was chosen over its counterpart based on the total amount of choices.
	Our SmartPatch method (\addlegendimageintext{draw=black,fill=color_ours,area legend}) outperforms both lineGen (\addlegendimageintext{draw=faublue,fill=color_lineGAN,area legend}) and GANwriting (\addlegendimageintext{draw=black,fill=color_GANwriting,area legend}) with pick rates of
	\SI{70.5}{\percent} and \SI{64.5}{\percent}. 
	Also, in \SI{54.4}{\percent} of the cases our results were more often chosen than the actual IAM images (\addlegendimageintext{draw=faublue,fill=color_real,area legend}).
	However, real data is favored with \SI{66.0}{\percent} over GANwriting and \SI{71.9}{\percent} over lineGen.
	In a direct comparing of the latter two methods, GANwriting is slightly preferred by \SI{53.5}{percent}. 
	
	\begin{figure}[t]
	    \centering
	    \begin{tikzpicture}
			\begin{customlegend}[%
				legend entries={Real data, GANwriting, lineGen, SmartPatch (ours)},
				legend style={%
					/tikz/every even column/.append	style={column sep=0.5cm}},
				legend columns=-1,
				legend cell align=left]
				\addlegendimage{draw=faublue,fill=color_real,area legend}
				\addlegendimage{draw=faublue,fill=color_GANwriting,area legend}
				\addlegendimage{draw=faublue,fill=color_lineGAN,area legend}
				\addlegendimage{draw=faublue,fill=color_ours,area legend}
			\end{customlegend}
		\end{tikzpicture}
		
		\vspace{2pt}
	    \centering
	    \includestandalone{figure/us_scatter}
	    \caption{Pick rate relative to word lengths. Small words have a length between 1 and 3 characters, medium words between 4 and 6 characters and large words between 7 and 10 characters.}
	    \label{fig:us_word_length}
	\end{figure}
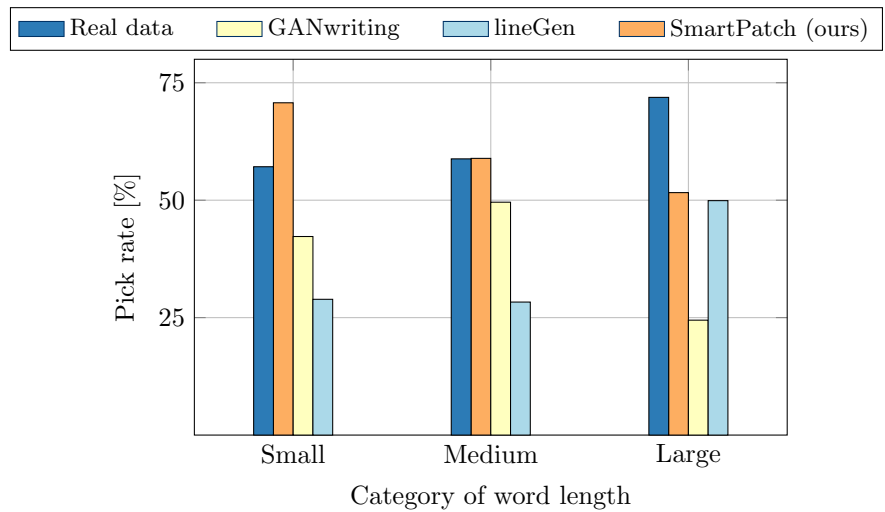
	\cref{fig:us_word_length} shows the different picking rates for different word lengths for each method. We categorized the words into small, medium, and large. 
	Small words contain one to three characters, medium words contain four to five characters and large words contain seven to 10 characters. 
	For small words, our method has the highest pick rate of \SI{70}{\percent}, even more than real images. Small words from GANwriting and lineGen are only chosen by \SI{42}{\percent} and \SI{28}{\percent}, respectively.
	For medium words, GANwriting improves to almost \SI{50}{\percent}, but lineGen stays nearly the same.
	Our approach performs equally well for medium words than real words, but for large words, the pick rate further increases for real data and our method is just above \SI{50}{\percent}. 
	GANwriting's pick rate decreases quite drastically to under \SI{25}{\percent} for large words whereas lineGen advances to almost \SI{50}{\percent}. 
	Also note, that for words with ten characters lineGen is picked in \SI{60}{\percent} of the cases.
	Our method performs the best for words of length three with a pick rate of \SI{72}{\percent}.

	\section{Discussion}
	\label{discussion}
	Having a higher pick rate than real images especially for small words is a very interesting outcome of the user study. 
	Presumably, the reason is that
	participants value readability significantly when comparing the ``realness'' of samples. 
	Especially small, hastily written words like ``the'' and ``to'' tend to be smudged by human writers, making them difficult to read individually. Our method penalizes unreadable characters explicitly, leading to more readable characters for small words (see \cref{fig:Userstudy}, top right). 
    This effect might have been enhanced by the fact that subjects had only a single word to analyze and were explicitly asked to distinguish real from machine-generated examples.
	Additionally, small artifacts introduced during the scanning of IAM samples or malfunctioning pens may also introduce artifacts that participants interpret as generation errors (see \cref{fig:Userstudy}, second row, first image).
	In these cases, the small inaccuracies in stroke consistency are interpreted not as errors in scanning or a dried-out pen, but rather as very weak versions of the grave errors witnessed \eg \cref{fig:bandingInGANwriting}.
	The evaluation of character recognition rate with the HTR system supports our thesis that GANwriting and our approach are focused on producing readable outputs. 
	Further, this questions the conducted user studies of all handwriting generation methods to evaluate whether the generated handwriting images are looking real or not. 
	Also only using HTR systems to grade will not be enough of a quantitative measurement if the improvements in this field further continue this fast. 
	Like Davis et al.~\cite{davis2020text} already note, we also doubt the validity of reporting the FID score as a good rating for different handwriting generation tasks.
	
	\section{Conclusion}
	\label{conclusion}
	\begin{figure}[t]
	    \centering
	    \captionsetup[subfigure]{labelformat=empty}
	    \begin{tikzpicture}
	        \begin{scope}[spy using outlines={rectangle,lens={scale=3},height=1.5cm,width=1cm, connect spies, ultra thick}, node distance=1cm]
	        \node[](base1){%
	            \includegraphics[width=0.2\textwidth]{images/comparisions/base/m04-123-02-05.png}
	        };
	        \node[right=\textwidth of base1.south west, anchor=south east](character1){%
	            \includegraphics[width=0.2\textwidth]{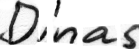}
	        };
	        \spy[orange,thick] on ($(base1)+(-0.85cm,0.2cm)$) in node (s1) [right=5cm of
            base1.north west,anchor=north east];
            \spy[orange,thick] on ($(character1)+(-1cm,0.0cm)$) in node (s2) [
            right = 2cm of s1
            ];
            \node[below=0.5cm of base1](base3){%
                \includegraphics[width=0.2\linewidth]{images/comparisions/base/m06-067-02-04.png}
            };
            \node[right=\textwidth of base3.south west, anchor=south east](character3){%
                \includegraphics[width=0.2\linewidth]{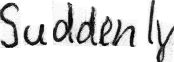}
            };
            \spy[orange,thick] on ($(base3)+(0.2cm,-0.1cm)$) in node (s5) [below=0.2 cm of s1];
            \spy[orange,thick] on ($(character3)+(0.3cm,-0.04cm)$) in node (s6) [below=0.2 cm of s2];
            
	        \end{scope}
	    \end{tikzpicture}
	    \caption{Comparisions between GANwriting (left) and our SmartPatch method (right).}
	    \label{fig:bandingInGANwriting_comparison}
	\end{figure}
	We introduced a new lightweight local patch discriminator that significantly improves the state-of-the-art system for handwritten word generation.
	Specifically, our method increases the verisimilitude of generated characters,
	as seen in \cref{fig:bandingInGANwriting_comparison}, by explicitly attending to characters that create more convincing outputs compared to the current state-of-the-art.
	This extra feedback module was developed especially for this use case but shows the great improvement of incorporating prior knowledge into the training process.  
	We performed a user study that further shows the use of our method for generating handwritten words and that it produces more realistic text than other state-of-art methods.

	\appendix
    \bibliographystyle{splncs04}
    \bibliography{Ganwriting}
    
    \renewcommand\thesection{\Alph{section}}

    \section*{APPENDIX}
    \section{Further examples}
    \label{appendix:examples}
    \begin{figure}
        \centering
        \includestandalone[width=0.95\textwidth]{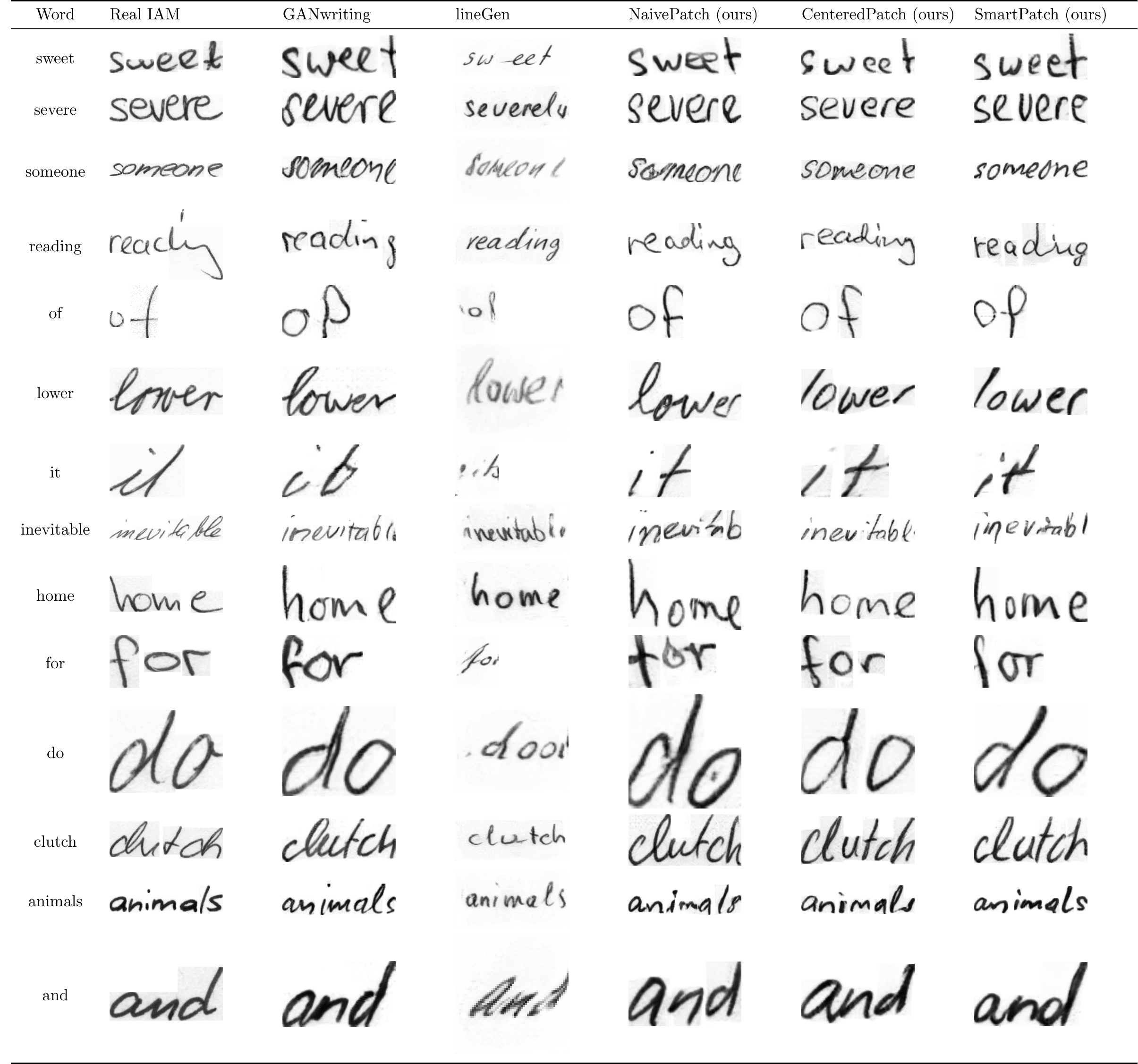}
        \caption{Comparison of randomly chosen outputs. For each row the priming image and the content is the same.}
        \label{fig:qualitative_appendix}
    \end{figure}

\end{document}